\documentclass{bmvc2k}

\usepackage{wrapfig,booktabs}
\usepackage{pifont}% http://ctan.org/pkg/pifont
\usepackage{algorithm}
\usepackage{algpseudocode}
\usepackage{booktabs}       % professional-quality tables
\usepackage{multirow}
\usepackage{bm}
\usepackage{color}
\usepackage{colortbl}
\usepackage{enumitem}
\usepackage{pifont}  
\usepackage{mathtools}
\usepackage{array}
\usepackage{enumitem}
\usepackage{pifont}
\usepackage{sidecap}
\usepackage{xspace}
\usepackage{listings}
\usepackage{graphicx}

\newcommand{\dec}[1]{\ensuremath{_{\text{\textcolor{magenta}{(-#1)}}}}}

\definecolor{Gray}{gray}{0.90}
\definecolor{white}{rgb}{1.0, 1.0, 1.0}

\definecolor{LightCyan}{RGB}{240, 224, 238}
\newcolumntype{a}{>{\columncolor{LightCyan}}c}

\title{Adversarial Pixel Restoration as a Pretext Task for Transferable Perturbations}

\addauthor{Hashmat Shadab Malik}{hashmat.malik@mbzuai.ac.ae}{1}
\addauthor{Shahina K Kunhimon}{shahina.kunhimon@mbzuai.ac.ae}{1}
\addauthor{Muzammal Naseer}{muzammal.naseer@mbzuai.ac.ae}{ 1,2}
\addauthor{Salman Khan}{salman.khan@mbzuai.ac.ae}{ 1,2}
\addauthor{Fahad Shahbaz Khan}{fahad.khan@mbzuai.ac.ae}{ 1,3}

\addinstitution{
 Mohamed bin Zayed University of AI\\
}
\addinstitution{
 Australian National University
}
\addinstitution{
 Link\"oping University
}
\runninghead{Malik et.al}{Adversarial Pixel Restoration}%{Adv. Pixel Rest. Task for Transferable Perturbations}

\def\eg{\emph{e.g}\bmvaOneDot}

\def\etal{\emph{et al}\bmvaOneDot}
\begin{document}

\maketitle
\begin{abstract}
Transferable adversarial attacks optimize adversaries from a pretrained surrogate model and known label space to fool the unknown black-box models. Therefore, these attacks are restricted by the availability of an effective surrogate model. In this work, we relax this assumption and propose \emph{Adversarial Pixel Restoration} as a self-supervised alternative to train an effective surrogate model from scratch  under the condition of \emph{no} labels and \emph{few} data samples. Our training approach is based on a min-max scheme which reduces overfitting via an adversarial objective and thus optimizes for a more generalizable surrogate model. Our proposed attack is complimentary to the adversarial pixel restoration and is independent of any task specific objective as it can be launched in a self-supervised manner.  We successfully demonstrate the adversarial transferability of our approach to Vision Transformers as well as Convolutional Neural Networks for the tasks of classification, object detection, and video segmentation.  Our training approach improves the transferability of the baseline unsupervised training method 
by  16.4\% on ImageNet val. set. Our codes \& pre-trained surrogate models are available at: \href{https://github.com/HashmatShadab/APR}{https://github.com/HashmatShadab/APR}.
\end{abstract}

%------------------------------------------------------------------------- 
\section{Introduction}
Adversarial attacks \cite{Szegedy2014, Goodfellow2015, Papernot2016limitations, Moosavi2016, CW2017, Madry2018, Athalye2018obfuscated} add small, imperceptible but well optimized noise to the clean image which can elicit an incorrect decision from the model.
These attack methods  \cite{poursaeed2018generative, dong2017discovering, xie2019improving, inkawhich2020perturbing, gao2020patch, wang2021admix} craft adversarial examples that can broadly be categorized based on how much information is available about the target model. In a \emph{white-box} attack setting, the attacker has complete knowledge of the target model and can directly optimize adversarial perturbations for the given model. In a more realistic \emph{black-box} attack setting, the attacker does not have access to the target model, its architectural details or targeted task (\eg, classification, segmentation or object detection). In such a case, adversarial examples are created on a surrogate model and then transferred to the black-box model.  Adversarial examples generated from surrogate models trained on a large-scale dataset in a supervised manner have better transferability \cite{naseer2019cross, dong2018boosting}. Transferability of such attacks improves further by fine-tuning the surrogate model to enhance their representation capacity \eg, by finding better self-ensemble from a given pretrained model \cite{naseer2021improving}. The adversarial transferability hence depends on the generalizability of the surrogate model. Such attacks are also restricted by the availability of a pretrained surrogate and information about the label space.

In this work, we call into question this assumption and consider a stronger threat model where an attack is launched from few unannotated or cross-domain samples (\eg, painting to ImageNet samples) without any knowledge of target networks or tasks. This threat model poses a challenge on how to learn an effective surrogate model from the limited unlabelled data and then how to generate self-supervised transferable adversarial examples.  

With limited availability of data, neural networks can easily memorize the data \cite{zhang2021understanding} even with strong augmentations \cite{li2020practical}. Therefore, Li \etal \cite{li2020practical} propose to reconstruct transformed input images to learn a surrogate model. However, their effective surrogate training and attack approach requires supervision though annotated data samples. In order to reduce overfitting over few data samples and to find robust features, we take inspiration from adversarial training \cite{Goodfellow2015, springer2021little} and propose self-supervised Adversarial Pixel Restoration to train a surrogate model. The min-max objective of our proposed training allows to find a flatter minima with robust features which compliments our self-supervised adversarial attack to achieve higher adversarial transferability. Our main contributions are as follows:
\begin{itemize}
    \item We propose self-supervised Adversarial Pixel Restoration to find highly transferable patterns by learning over flatter loss surfaces. Our training approach allows launching cross-domain attacks without access to large-scale labeled data or pretrained models.
    \item Our adversarial attack is self-supervised in nature and  independent of any task-specific objective. For instance, our approach optimizes the robust transformed loss surface of the surrogate via fooling its reconstruction ability. This allows to generate task independent adversaries. Therefore, our approach can transfer perturbations to a variety of tasks as we demonstrate for classification, object detection, and segmentation.
    \item We provide a thorough analysis  of our proposed method to establish it's effectiveness. We observe that our approach leads to smoother loss landscapes (Fig. \ref{Loss surfaces}), helping in crafting more generalizable adversarial examples. Our method remains effective even in extreme data scarcity \eg when trained on two data samples only. (see Sec. \ref{subsec: Ablative Analysis}).
    
    % consistently performs well with varying training iterations and data samples.
    % and visualizations of our learned loss surfaces to establish the effectiveness of our proposed method.
\end{itemize}
 
\section{Related Work}

Several gradient-based methods  \cite{Goodfellow2015, kurakin2018adversarial, mlatscale, ensembleadv} have been proposed for crafting adversarial examples directly on the target classifiers. However, when the access to the target model is limited to just a finite amount of queries, current methods either rely on the transferability of surrogate models \cite{DasTdatafree, Papernot2016transferability, Papernot2017} or estimate the gradients/boundary of the target model \cite{ZOOgradest, decisonbased_Brendel2018, decison_based_Narodytska2017}. Both of the above approaches either require a non-trivial number of queries from the target model or access to the training distribution, making it highly impractical in real-case scenarios. A practical threat model was introduced in \cite{naseer2019cross}, where the attacker does not have access to the training distribution of the target model, as well as querying is prohibited. Authors train a generator-based surrogate model with the help of training data and a pretrained classifier obtained from a different domain than the target models. However, the pretrained classifier as well as the generator are trained on a large annotated dataset. 

In \cite{li2020practical}, a stronger threat model with access to limited data samples (order of tens) was proposed. Inspired from self-supervised learning methods, autoencoder-based surrogate models are trained with limited data. However, the transferability of the autoencoders trained in an unsupervised manner is still moderate. Most of the previous works have explored the transferability of surrogate  models trained on the target model's training set. In  \cite{springer2021adversarial}, the authors observe that the features of robust classifiers can be used to generate adversarial examples that are more transferable.  Building on this, \cite{springer2021little}  observe that having a classifier adversarially trained with a small perturbation budget (\emph{"slightly robust"}), leads to highly transferable adversarial examples.
Unlike prior works, we focus on constructing robust surrogate models in a fully unsupervised manner. We consider  the case of training robust model with limited data 
%with the goal 
to improve adversarial transferability. Furthermore, we also consider a practical scenario of availability of large unlabelled dataset for training of robust models which can be used for constructing cross-domain and cross-task  adversaries.%adversarial examples.
\section{Adversarial Pixel Restoration for Transferable Perturbations}
Our goal is to learn a surrogate model, $\mathcal{F}$, from a given unlabelled data distribution, $P_s$, with a set of only few image samples ($\leq 20$). This setting is in contrast to the existing transferable adversarial attacks \cite{dong2018boosting, naseer2021improving, xie2019improving, Naseer_2021_ICCV} that assume access to a surrogate model trained on a large-scale annotated data (\eg, ImageNet \cite{deng2009imagenet}) in a supervised fashion. In the presence of unlabeled data, however, the surrogate model can be trained by defining a self-supervised task, $\mathcal{T}_s$, such as predicting rotation \cite{gidaris2018unsupervised}, solving jigsaw puzzle \cite{noroozi2016unsupervised} or by matching different views of the same input sample \cite{caron2021emerging}. A major challenge is that deep neural networks can easily memorise the data \cite{zhang2021understanding} and quickly overfit the few available samples even after applying strong data augmentation techniques \cite{li2020practical}. This results in a surrogate model with less generalizable representations and consequently, adversarial attack launched from such a model has weak transferability (Sec.~\ref{subsec: Transferability from Few In-Domain Samples}).

\begin{figure}[t]
\centering
\includegraphics[width=\textwidth]{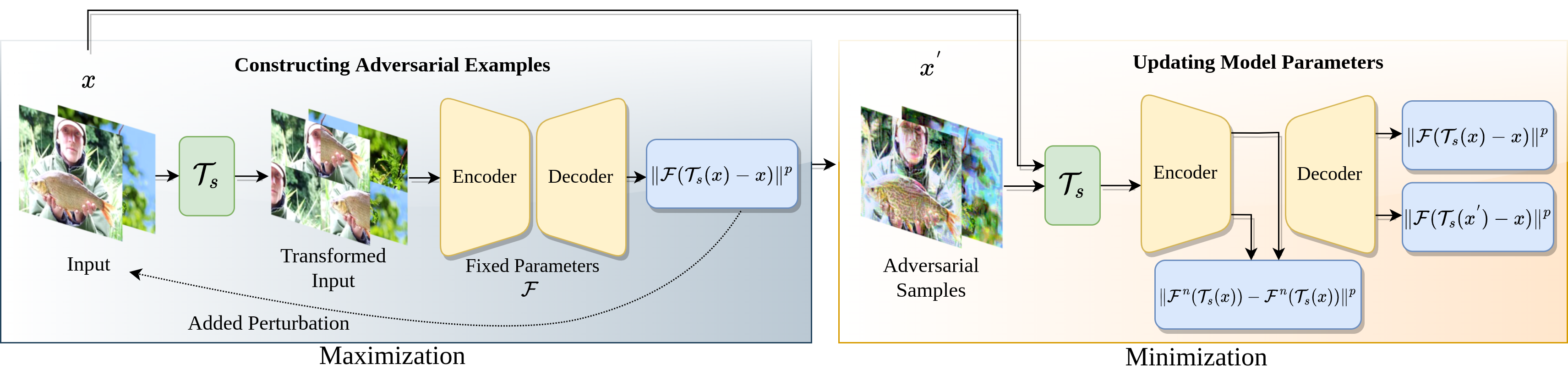} \vspace{-0.7em}
\vspace{-2em}
\caption{\small Our approach trains an autoencoder based surrogate model via self-supervised adversarial pixel restoration to learn generalizable representations from a limited number of data samples ($\le 20$). Our training is based on a min-max strategy. We first generate adversarial examples by fooling model's reconstructive ability \emph{(maximization)}, followed by updating the model parameters based on restoration of the transformed adversarial and clean sample \emph{(minimization)}. Our approach allows launching transferable self-supervised adversarial attacks without any knowledge of target (black-box) model.
}\vspace{-1em}
\label{fig: concept_fig}
\end{figure}

We propose adversarial pixel restoration as a prior to train the surrogate model $\mathcal{F}$ that boosts transferability of the adversarial attacks (see Fig. \ref{fig: concept_fig} ). We create adversarial examples from the original input images by attacking the source model in the pixel space using our proposed adversarial attack (Sec.~\ref{subsec: Adversarial Pixel Transformations}). Our approach also shifts the position of input pixels via transformations such as rotation or jigsaw shuffle. We then train a model to denoise and restore pixels to their original positions via our proposed adversarial training (Algo. \ref{algo: Adversarial Denoising and Restoration}).   The surrogate model $\mathcal{F}$ in our case consists of an autoencoder as explained below.

\subsection{Surrogate Architecture} 
\label{subsec: Surrogate Architecture}
 The surrogate model is based on an autoencoder \cite{li2020practical}. The encoder consists of a stack of convolution layers. At the beginning of the architecture, convolution layers with a larger kernel size and stride help reduce the spatial resolution of the feature maps, followed by multiple residual blocks where the size of the feature maps is kept constant. The decoder is a lightweight model consisting of two transpose convolution layers to upsample the feature maps. The adversarial training of these autoencoders is based on denoising and pixel restoration as explained next in Sec.~\ref{subsec: Adversarial Training via Denoising}.
 
\subsection{Adversarial Training via Denoising }
\label{subsec: Adversarial Training via Denoising}
\vspace{-0.3em}
The feature space of a slightly robust classifier produces highly transferable adversarial examples \cite{springer2021little}. However, adversarial training of such models is computationally demanding due to iterative training and also requires a large-scale labeled dataset that might not be accessible to the attacker in real-world scenarios. We assume a more practical threat model, where the attacker has access to a data distribution $P_s$ with limited number of samples without any annotations. We adversarially learn a surrogate model on the data distribution $P_s$ to transfer adversarial perturbations to the target (black-box) models trained for different tasks (\eg, classifications, object detection and segmentation) on possibly different target distribution $Q_t\neq P_s$. In such challenging attack settings, we adversarially train an autoencoder $\mathcal{F}$ via min-max training strategy \cite{Goodfellow2015}. At maximization step, we create adversarial examples by fooling the model $\mathcal{F}$ via adversarial pixel transformations with a single step attack (Algo. \ref{algo: Adversarial Denoising and Restoration}). At minimization step, we denoise and restore the feature and pixel space to achieve generalizable loss surfaces (Fig. \ref{Loss surfaces}) which leads to more transferable attack.

\begin{algorithm}[t]
	\caption{
		Adversarial Denoising and Restoration
	}
	\label{algo: Adversarial Denoising and Restoration}
	\begin{algorithmic}[1]
		\Require Source data distribution $P_s$, pixel transformation $\mathcal{T}_s$, attack step size $\delta$, perceptual budget $\epsilon$, balancing paramter $\lambda$, and maximum training iterations $T$
		\Ensure Randomly initialize surrogate model $\mathcal{F}$
		\For {$t\in [1, 2, \ldots T]$}
    		\State Randomly sample from $P_s$: $\bm{x} \sim P_s$
    		\State Initialize adversary $\bm{x}' \leftarrow$ $\bm{x}$
    		\State Optimize  $\bm{x}'$ using Eq. \ref{eq: attack_loss} or \ref{eq: attack_loss_prototypical}: \algorithmiccomment{Adversarial Pixel Transformation}
    		\begin{equation*}
    		    \bm{x}' \gets \bm{x} + \delta\times\text{sign}\left(\nabla\mathcal{L}_{max}\right)
    		\end{equation*}
    		
    		\State Project adversaries within allowed perceptual budget: $\bm{x}' \gets \text{clip}\left(\bm{x}', \bm{x}-\epsilon, \bm{x}+\epsilon \right)$
    		
    		\State Forward-pass $\bm{x}$ and $\bm{x}'$ through model $\mathcal{F}$ and update its parameters $\theta$ by minimizing the loss given in Eq. \ref{eq: overall_loss}: \algorithmiccomment{Pixel Restoration}
    		\begin{equation*}
    		    \theta \gets \theta - \alpha\times\nabla\mathcal{L}_{min},
    		\end{equation*}
    		where $\alpha$ is the learning rate.
		\EndFor
	\end{algorithmic}
\end{algorithm}

\subsubsection{Adversarial Pixel Transformations}
\label{subsec: Adversarial Pixel Transformations}
For given input samples $\bm{x} \sim P_s$, we first find adversarial example, $\bm{x}' \;\; \text{subjected to:} \;\;  \Vert \bm{x} - \bm{x}' \Vert_{\infty} \leq \epsilon$, by maximizing the following objective ($\mathcal{L}_{max}$):
\begin{align}
\label{eq: attack_loss}
\underset{\bm{x}'}{\text{maximize}} \quad \mathcal{L}_{max} =  \|\mathcal{F} \left(\mathcal{T}_s(\bm{x}')\right)- \bm{x}\|^p,
\end{align}
where $\mathcal{T}_s$ represents the pixel transformation (\eg, rotation or jigsaw shuffle) that shifts the pixel positions. Therefore, our attack fools the model's ability to restore the transformed pixel space by maximizing the loss presented in Eq.~\ref{eq: attack_loss} and ultimately help to robustify self-supervised features. Our attack approach can also be extended to benefit from supervisory signals \eg, by fooling prototypes \cite{snell2017prototypical, li2020practical} as follows:
\begin{equation}
\label{eq: attack_loss_prototypical}
\underset{\bm{x}'}{\text{maximize}} \quad \sum^{C}_{c=1} \left(y_c \, \|\mathcal{F} \left(\mathcal{T}_s(\bm{x}')\right)- \bm{x}^{(c)}\|^p\right),
\end{equation}
where $C$ represents the number of categories, $y_c$ represents one-hot encoded labels, and $\bm{x}^{(c)}$ is the chosen prototype for a particular class. Our proposed attack objective in Eq.~\ref{eq: attack_loss_prototypical} helps to optimize for robust discriminative features with better adversarial transferability.

\subsubsection{Pixel Restoration}
\label{pixel_restoration}
For a given adversarial sample, $\bm{x}'$ created using Eqs. \ref{eq: attack_loss} or \ref{eq: attack_loss_prototypical},  we train the surrogate model, $\mathcal{F}$, by pixel restoration. Our loss function penalizes the model $\mathcal{F}$ by minimizing the reconstruction error between the original sample, $\bm{x}$, and the model's output for the transformed adversarial as well as the transformed original sample as follows:
\begin{equation}
\label{eq: out_loss}
    \mathcal{L}_{out} =\|\mathcal{F} \left(\mathcal{T}_s(\bm{x}')\right)-  \bm{x}\|^p + \|\mathcal{F} \left(\mathcal{T}_s(\bm{x})\right)-  \bm{x}\|^p.
\end{equation}

We further regulate the model's feature space during adversarial training by enforcing alignment between the original and adversarial feature distributions as follows:
\begin{equation}
\label{eq: feature_loss}
\mathcal{L}_{{feature}}=\|\mathcal{F}^{n} \left(\mathcal{T}_s(\bm{x}')\right)- \mathcal{F}^{n} \left(\mathcal{T}_s(\bm{x})\right)\|^p,
\end{equation}
where $\mathcal{F}^{n}$ represents the intermediate (encoder) layer output. Overall training objective is,
\begin{equation}
\label{eq: overall_loss}
    \mathcal{L}_{min} = \mathcal{L}_{out} + \lambda \mathcal{L}_{feature},
\end{equation}
where $\lambda$ is the balancing parameter.

\subsubsection{Behavior of Robust Loss Surfaces}
\begin{figure}[t]
\begin{minipage}{0.69\linewidth}
    \centering
     % lelf lower right up 
    \includegraphics[trim= 20mm 30mm 20mm 40mm, clip, height=2.9cm, width=0.49\textwidth]{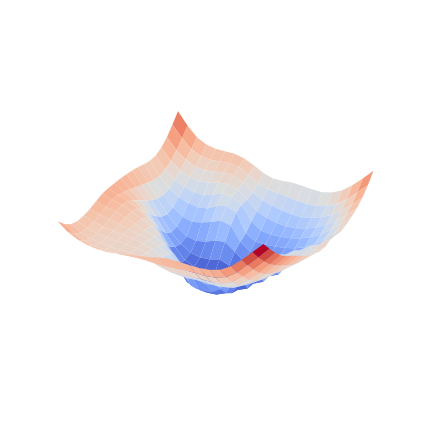}
    \includegraphics[trim= 20mm 30mm 20mm 40mm, clip,  width=0.49\textwidth]{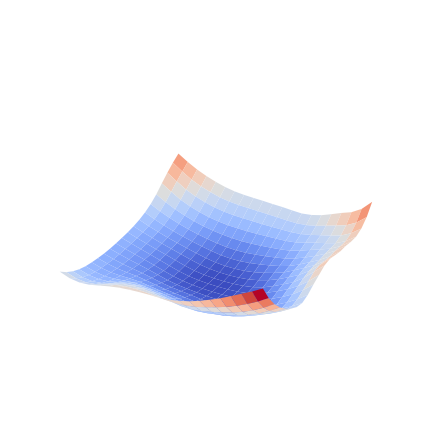}\\
    \vspace{-1em}
     \includegraphics[trim= 20mm 30mm 20mm 50mm, clip, width=0.49\textwidth]{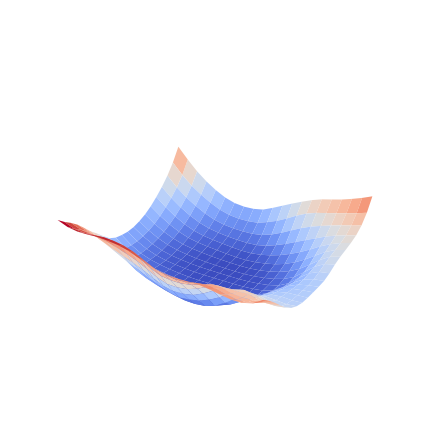}
    \includegraphics[trim= 20mm 30mm 20mm 50mm, clip, height=2.5cm, width=0.49\textwidth]{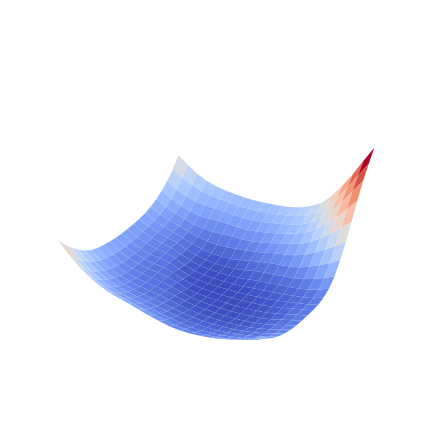}
\end{minipage}
\hspace{0.01\linewidth}
\begin{minipage}{0.28\linewidth}
    \caption{\small Reconstruction loss landscape of autoencoders. The first column shows the loss landscapes of autoencoders trained using the baseline method on the task of rotation (\emph{top}) and jigsaw (\emph{bottom}). The second column shows the corresponding flatter loss surfaces obtained by training the autoencoders using our approach.}
    \label{Loss surfaces}
\end{minipage}
\end{figure}

We visualize the loss landscapes of our trained autoencoders (Fig. \ref{Loss surfaces}). We use the filter normalization method proposed in \cite{li2018visualizing}, which shows the structure of the loss surface along with random directions near the optimal pretrained parameters. We observe that our approach leads to more flatter minima as compared to the baseline method \cite{li2020practical}. This has significant effect on finding generalizable adversarial example with better transferability (Sec.~\ref{subsec: Transferability from Few In-Domain Samples}).

 \section{Experimental Protocols}
 \label{sec: Experimental Protocols}

For a pixel range [0, 1], we create $l_\infty$ adversarial examples under perceptual budget of $\epsilon \le 0.1$ following \cite{li2020practical}. We show adversarial transferability of our approach against ImageNet trained models (Convolutional and ViTs), object detection and segmentation from in-domain and cross-domain (paintings, medical scans ) data samples. We used Adam \cite{kingma2014adam} optimizer with a learning rate of 0.001 for our proposed adversarial pixel restoration (Algo. \ref{algo: Adversarial Denoising and Restoration}) at $\epsilon \le \frac{2}{255}$, $\delta = \frac{2}{255}$, $\lambda=1$, $p=2$. We provide detailed ablative analysis on the effect of these hyper-parameters in Sec.~\ref{subsec: Ablative Analysis}.

\noindent\textbf{Surrogate Training  with Few Samples:} Similar to \cite{li2020practical}, we assume access to only few data samples (\eg, 20 samples to train a single autoencoder). We apply Eq. \ref{eq: attack_loss} as adversarial pixel transformation ($\mathcal{T}$) based on rotation or jigsaw in an unsupervised setting. We also incorporate our method in the supervised prototypical training of autoencoders mentioned in \cite{li2020practical}, where the reconstruction objective function used during the minimization and maximization step is formulated by Eq. \ref{eq: attack_loss_prototypical}. We compare the best prototypical setting reported in \cite{li2020practical}, comprising of multiple decoder networks. All models are trained for the same no. of iterations as \cite{li2020practical}. We provide pseudo code of Algo. \ref{algo: Adversarial Denoising and Restoration} in Appendix  \ref{appendix: pseudocode}.

\noindent\textbf{Surrogate Training with Large Dataset:}  There is abundance of unannotated data available via online sources. Therefore, we also scale our proposed self-supervised adversarial training to large-scale datasets. Specifically, we train a surrogate model (single autoencoder) on paintings (79k samples) \cite{paintings}, CoCo (40k samples) \cite{lin2014microsoft}, and Comics (50k samples) \cite{comics} to test unsupervised cross-domain adversarial transferability of our method.   

\noindent\textbf{Target Models:}
Adversarial perturbations from our trained autoencoders (surrogate models) are transferred to classification models including convolutional: VGG-19 \cite{Simonyan2015}, Inception-v3 (Inc-V3) \cite{Szegedy2016}, ResNet-152 (Res152) \cite{He2016}, Dense161 \cite{Huang2017densely}, SeNet \cite{Hu2018}, Wide-ResNet-50 (WRN) \cite{Zagoruyko2016}, and MobileNet-V2 (MNet-V2) \cite{Sandler2018mobilenetv2}, and Vision Transformers: ViT-T and ViT-S \cite{dosovitskiy2020image}, DeiT-T and DeiT-S \cite{touvron2021training}. We also evaluated adversarial vulnerability of robust ResNet-50 models \cite{salman2020adversarially}. Further, we transfer attack to DETR \cite{carion2020end} and DINO \cite{caron2021emerging} to evaluate on object detection and video segmentation tasks. 

\noindent\textbf{Evaluation Metrics:} We report drop in Top-1 (\%) accuracy, Mean Average Precision (mAP), and Jaccard Index for classification, object detection and segmentation, respectively. 

\noindent\textbf{Datasets:} We evaluate classification models on 5k samples from ImageNet validation set in the same setting as \cite{li2020practical}. DETR and DINO are evaluated on CoCo (5k samples) and DAVIS (2k samples) validation sets respectively.

\noindent\textbf{Baseline Adversarial Attack:} For adversarial prototypical training, we use the same supervised attack objective as proposed by \cite{li2020practical} for direct comparison. Specifically, we use 200 iterations of I-FGSM \cite{kurakin2018adversarial} followed by 100 iterations of ILA \cite{huang2019enhancing}. The attack objective for the surrogate models trained in self-supervised manner is simply based on maximizing the reconstruction error as described in Eq. \ref{eq: attack_loss}.

\subsection{Results}

\subsubsection{Transferability from Few In-Domain Samples}
\label{subsec: Transferability from Few In-Domain Samples}
\begin{figure}[!t]
\small \centering
\begin{minipage}{0.7\textwidth}
    \begin{minipage}{.49\textwidth}
        \centering
        \includegraphics[width=\linewidth]{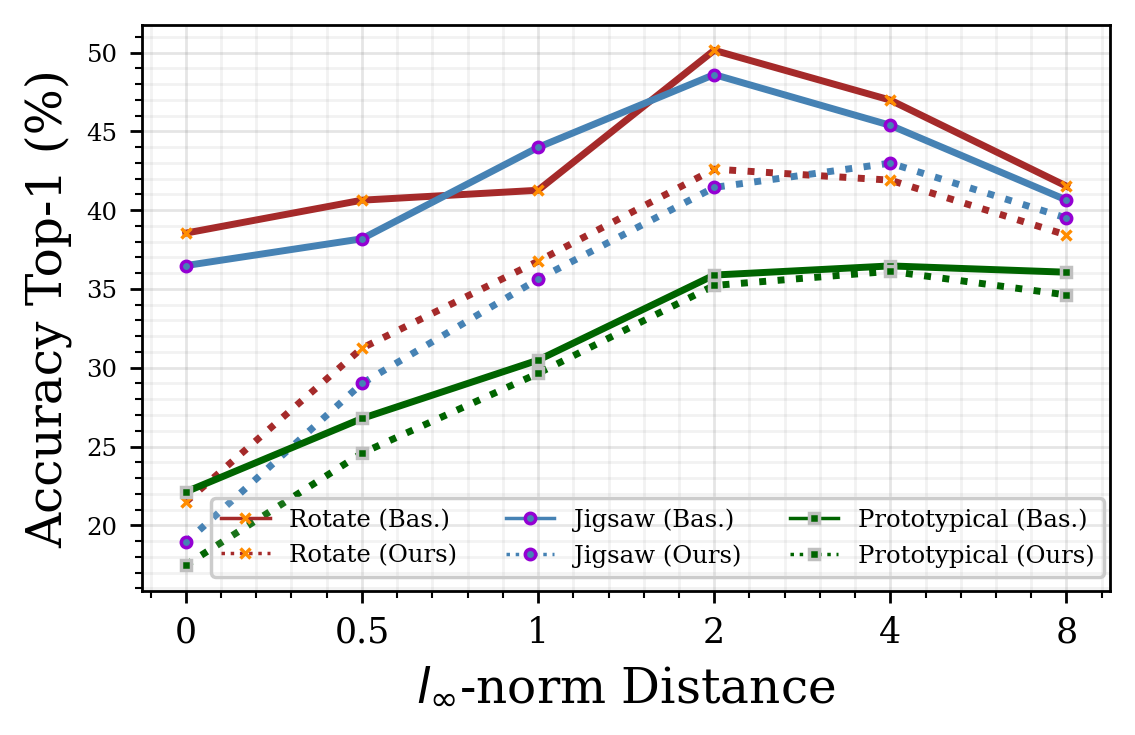}
    \end{minipage}
    \begin{minipage}{.49\textwidth}
        \centering
        \includegraphics[width=\linewidth]{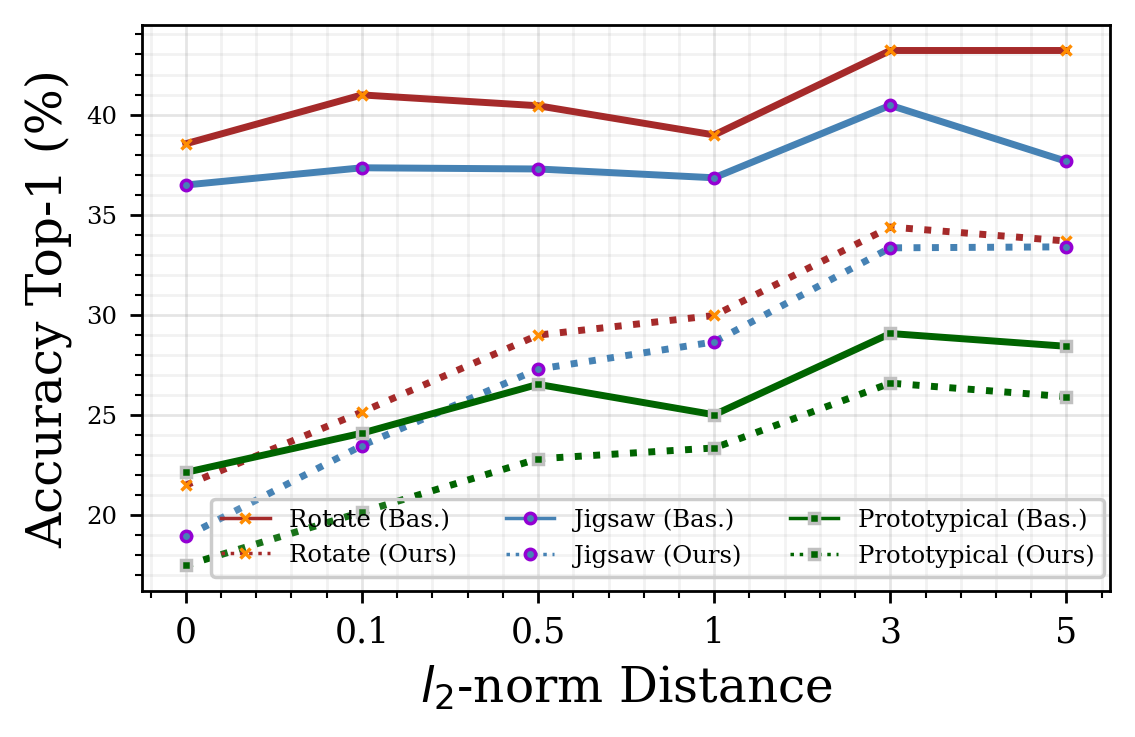} 
	\end{minipage}
\end{minipage}
\begin{minipage}{0.28\textwidth}
\vspace{-1em}
    \caption{\small Models trained on $l_\infty$ examples and large norm distance are less vulnerable to our attack, however, such models also lose accuracy on clean images \cite{salman2020adversarially}.}
\label{fig: transferability against robust models}
\end{minipage}
\vspace{-0.5em}
\end{figure}

Our surrogate models trained on few data samples show significantly higher adversarial transferability as compared to the baseline \cite{li2020practical}. Attack generated on our self-supervised models (rotation, and jigsaw) performs even better than supervised (prototypical) models of \cite{li2020practical} (Tables \ref{tab:no-box table}, \ref{tab:no-box table_transformers}). Our approach further boosts the transferability rates when combined with supervised adversarial prototypical training. This compliments the benefits of our method in both supervised and self-supervised settings. We observe that vision transformers are more robust as compared to convolutional networks against such attacks \cite{naseer2021intriguing}, however, our approach provides non-trivial gains in fooling the Vision Transformers (Table \ref{tab:no-box table_transformers}). Similarly, adversarially robust models \cite{salman2020adversarially} are less vulnerable to such attacks (Fig. \ref{fig: transferability against robust models}).

\begin{table}[!t]
\centering\small
   \scalebox{0.8}[0.8]{
    \begin{tabular}{lccccccccl}
        \toprule
        \rowcolor{Gray} 
        Transformation & Method  & VGG-19  & Inc-V3  & Res152  & Dense121  &  SeNet & WRN  &  MNet-V2  & Average \\ \midrule
        \multirow{2}{*}{\rotatebox[origin=c]{0}{Jigsaw}} & \cite{li2020practical}  & 31.54 &  50.28 &  46.24 & 42.38 & 59.06& 51.24  & 25.24 & 43.71 \\ 
         & Ours   & 16.82 &  25.54 & 31.18 & 22.64 & 38.06  & 25.76 & 13.70 & \textbf{24.81}\dec{18.9}\\ 
         \midrule
        \multirow{2}{*}{\rotatebox[origin=c]{0}{Rotation}} & \cite{li2020practical}   & 31.14 &  48.14 & 47.40  & 41.26 & 58.20& 50.72 & 26.00 & 43.27\\
        & Ours   & 19.02 & 25.76  & 33.60 & 25.60 & 38.92  & 29.78 & 15.38 & \textbf{26.87}\dec{16.4}\\
        \midrule
        \multirow{2}{*}{\rotatebox[origin=c]{0}{Prototypical}}  & \cite{li2020practical} & 18.74 &  33.68& 34.72 & 26.06 & 42.36 & 33.14  & 16.34 & 29.29\\
        &  Ours 	 &17.02  & 21.48  & 28.66 & 21.06 & 35.04  & 23.56 & 13.06& \textbf{22.84}\dec{6.45}
        \\\midrule
        % \multicolumn{9}{l}{\footnotesize{$^\ast$ }}\\
    \end{tabular}}
    \vspace{0.3em}
    \caption{\small Our attack consistently boost adversarial transferability across ImageNet trained models. Our training approach (Algo. \ref{algo: Adversarial Denoising and Restoration}) hence prove to be complimentary to autoencoders trained with different self-supervised (SS) tasks. Results (Top-1 (\%), \emph{lower is better})  are reported on 5k images from ImageNet validation set introduced by \cite{li2020practical} under the same perceptual budget ($\epsilon \le 0.1$). }
    \label{tab:no-box table}
    % \vspace{-0.5em}
\end{table}

\begin{table}[!t]
    \begin{minipage}{.55\textwidth}
        \centering\small
        \setlength{\tabcolsep}{5pt}
        
        \scalebox{0.8}[0.8]{
    \begin{tabular}{lcccccl}
        \toprule
        \rowcolor{Gray} 
         Transformation & Method  & Deit-T  & Deit-S  & ViT-T  & ViT-S  & Average \\ \midrule
        \multirow{2}{*}{\rotatebox[origin=c]{0}{Jigsaw}} & \cite{li2020practical}  & 45.3 &  62.42 &  36.62 & 62.1 & 51.61 \\ 
         & Ours   & 43.50 &  59.50 & 18.0 & 52.48 & \textbf{43.37}\dec{8.24}\\ 
         \midrule
        \multirow{2}{*}{\rotatebox[origin=c]{0}{Rotation}} & \cite{li2020practical}   & 46.1 &  62.0 & 37.8  & 60.38 & 51.57\\
        & Ours   & 40.84 & 55.22  & 19.64 & 48.62 & \textbf{41.08}\dec{10.49}\\
        \midrule
        \multirow{2}{*}{\rotatebox[origin=c]{0}{Prototypical}}  & \cite{li2020practical} & 38.18 &  54.96& 21.5 & 50.28 & 41.23\\
        &  Ours 	 &34.16  & 51.54  & 16.74 & 45.3 & \textbf{36.94}\dec{4.29}  
        \\\midrule
        % \multicolumn{9}{l}{\footnotesize{$^\ast$ }}\\
    \end{tabular}}
    \end{minipage}
    \hfill
   \begin{minipage}{.34\textwidth}
   \vspace{-1em}
    \caption{\small Comparative analysis of adversarial transferability for Vision  Transformers models on ImageNet validation set. The Top-1 (\%) under $l_\infty$ bound $\epsilon=0.1$  is shown (\emph{lower is better}). Our method performs favorably well.}
    \label{tab:no-box table_transformers}
    \end{minipage} 
    \vspace{-0.5em}
\end{table}

% \begin{SCtable}[][!t]
% \centering\small
% %   \setlength{\tabcolsep}{1.8pt}
%   \scalebox{0.72}[0.72]{
%     \begin{tabular}{lcccccl}
%         \toprule
%         \rowcolor{Gray} 
%          Transformation & Method  & Deit-T  & Deit-S  & ViT-T  & ViT-S  & Average \\ \midrule
%         \multirow{2}{*}{\rotatebox[origin=c]{0}{Jigsaw}} & \cite{li2020practical}  & 45.3 &  62.42 &  36.62 & 62.1 & 51.61 \\ 
%          & Ours   & 43.5 &  59.5 & 18 & 52.48 & \textbf{43.37}\dec{8.24}\\ 
%          \midrule
%         \multirow{2}{*}{\rotatebox[origin=c]{0}{Rotation}} & \cite{li2020practical}   & 46.1 &  62 & 37.8  & 60.38 & 51.57\\
%         & Ours   & 40.84 & 55.22  & 19.64 & 48.62 & \textbf{41.08}\dec{10.49}\\
%         \midrule
%         \multirow{2}{*}{\rotatebox[origin=c]{0}{Prototypical}}  & \cite{li2020practical} & 38.18 &  54.96& 21.5 & 50.28 & 41.23\\
%         &  Ours 	 &34.16  & 51.54  & 16.74 & 45.3 & \textbf{36.94}\dec{4.29}  
%         \\\midrule
%         % \multicolumn{9}{l}{\footnotesize{$^\ast$ }}\\
%     \end{tabular}}
%     \caption{\small Compare the transferability of adversarial examples crafted on transformer-based models on ImageNet. The Top-1 (\%) on adversarial examples under $l_\infty$ bound $\epsilon=0.1$  is shown (\emph{lower is better}).}
%     \label{tab:no-box table_transformers}
% \end{SCtable}
\begin{table}[t]
    \begin{minipage}{.44\textwidth}
        \centering\small
        \setlength{\tabcolsep}{4pt}
        
        \scalebox{0.8}[0.8]{
        \begin{tabular}{clca}
        \toprule 
            \rowcolor{Gray}
          Transformation & Method ($\rightarrow$) & {\cite{li2020practical}}& {Ours}  \\
        \midrule
        
        \multirow{3}{*}{\rotatebox[origin=c]{0}{Rotation}} & CoCo & 28.56 & \textbf{23.31} \\
        & Paintings & 27.83 & \textbf{17.75 } \\ 
        & Comics & 58.38 & \textbf{24.19 } \\ 
        \midrule
        \multirow{3}{*}{\rotatebox[origin=c]{0}{Jigsaw}} & CoCo & 43.93 & \textbf{31.28} \\
        & Paintings&  44.07 & \textbf{33.42}\\ 
        & Comics & 67.70 & \textbf{41.54 } \\ 
        \bottomrule
        \end{tabular}
        }
    \end{minipage}
    \hfill
   \begin{minipage}{.55\textwidth}
    \caption{\small Adversarial perturbations are transferred from a  single auto-encoder trained on CoCo, Painintgs or Comics. We report average of Top-1 (\%) accuracy against targeted conovlutional networks (Table \ref{tab:no-box table}). Our proposed approach provide favorable results (\emph{lower is better}). }
    \label{tab: single_surrogate_classification_results}
    \end{minipage} 
\end{table}
\begin{table}[!t]
    \begin{minipage}{0.49\textwidth}
    \centering\small
        \setlength{\tabcolsep}{3pt}
        \scalebox{0.6}[0.6]{
        \begin{tabular}{c|c|ca|ca|ca}
        \toprule 
        \rowcolor{Gray}

          Transformation ($\downarrow$) & \multirow{1}{*}{Dataset ($\rightarrow$)} & \multicolumn{2}{c|}{CoCo}& \multicolumn{2}{c}{Paintings} & \multicolumn{2}{c}{Comics}\\
        \midrule
        \rowcolor{white}
         & \texttt{No Attack} & \cite{li2020practical} & \texttt{Ours} & \cite{li2020practical} & \texttt{Ours} & \cite{li2020practical} & \texttt{Ours}\\
         \cline{2-6}
        
        Rotation & 39.7 & 19.3 & \textbf{14.6}  & 17.2 & \textbf{11.9} & 34.3 & \textbf{13.3} \\
        \midrule
        Jigsaw & 39.7 & 24.7 & \textbf{14.5}  & 24.1 & \textbf{14} & 38 & \textbf{20.8} \\

        \bottomrule
        \end{tabular}}
        \vspace{0.5em}
        \caption{\small Adversarial transferability to object detector (DETR) based on mAP at [0.5:0.95] is evaluated on CoCo validation set \cite{lin2014microsoft}.}
        \label{tab:DETR_results}
        
    \end{minipage}
    \hfill
    \begin{minipage}{0.49\textwidth}
        \centering\small
        \setlength{\tabcolsep}{3pt}
        \scalebox{0.6}[0.6]{
        \begin{tabular}{c|c|ca|ca|ca}
            \toprule 
            \rowcolor{Gray}
              Transformation ($\downarrow$) & \multirow{1}{*}{Dataset ($\rightarrow$)} & \multicolumn{2}{c|}{CoCo}& \multicolumn{2}{c}{Paintings}& \multicolumn{2}{c}{Comics}\\
            \midrule
            \rowcolor{white}
             &\texttt{No Attack} & \cite{li2020practical} & \texttt{Ours} & \cite{li2020practical} & \texttt{Ours} & \cite{li2020practical} & \texttt{Ours}\\
            \cline{2-6}
            
            Rotation &61.8 & 53.2 & \textbf{48.9}  & 52.6 & \textbf{46.9} & 57.78 & \textbf{47.81}  \\
            \midrule
            Jigsaw & 61.8 & 53.9 & \textbf{46.6} & 53.2 & \textbf{48.5} & 58.29 & \textbf{51.65}\\
            \midrule
            \bottomrule
            \end{tabular}
            }
            \vspace{0.5em}
            \caption{\small Adversarial transferability to object segmentation (DINO) based on Jacard index metric is evaluated on DAVIS validation set \cite{pont20172017}. }
            \label{tab:DINO_results}
        
    \end{minipage}
    
\end{table}

\subsubsection{Transferability from Cross-Domain Samples}
Extra unsupervised data can boost the performance of adversarial training \cite{carmon2019unlabeled}. We extend our approach to large-scale, unlabelled datasets to observe its effect on adversarial transferability of our approach as explained below. All surrogate models are trained for 50 epochs for cross-task adversarial transferability experiments.

% \subsubsection{Classification}

\noindent \textbf{Classification:} In this task, we train a single autoencoder and transfer its adversarial perturbations to ImageNet trained convolutional models as described in Sec.~\ref{tab: single_surrogate_classification_results}. We observe that our proposed adversarial training significantly improves upon the baseline \cite{li2020practical} (Table \ref{tab: single_surrogate_classification_results}).  We further note that the surrogate models trained on `paintings' dataset show higher adversarial transferability while `Rotation' as pixel transformation performs better. Further analysis on adversarial transferability of our attack is provided in Appendix \ref{appendix: results_eps_0.08}.

% \subsubsection{Object Detection and Segmentation}
\noindent \textbf{Object Detection and Segmentation:} Our adversarial attack based on simple transformed reconstruction error (Eq.~\ref{eq: attack_loss}) compliments our proposed adversarial training and successfully fools DETR for object detection and DINO for video segmentation (Tables \ref{tab:DETR_results} \& \ref{tab:DINO_results}). This signifies that our attack can be launched in real-world setting without any knowledge about the deployed vision system.

\subsection{Ablative Analysis}
\label{subsec: Ablative Analysis}
We thoroughly analyze and develop better understanding about the behavior of our approach by studying the effect of its different components including, \textbf{a)} \emph{Effect of Adversarial Pixel Restoration Prior on Training}, \textbf{b)} \emph{Effect of Perceptual Budget $\epsilon$ for our Single step Attack (Eq. \ref{eq: attack_loss})}, \textbf{c)} \emph{Effect of Iterative 
Attack during Training}, \textbf{d)} \emph{Effect of Training Iterations and Data size}, and \textbf{e)} \emph{Contribution of Losses}. All ablative experiments are conducted in limited data setting (Sec.~\ref{subsec: Transferability from Few In-Domain Samples}).

\begin{figure}[!t]
\small \centering
\begin{minipage}{0.68\linewidth}
    \begin{minipage}{.49\linewidth}
        \centering
        \includegraphics[width=\linewidth]{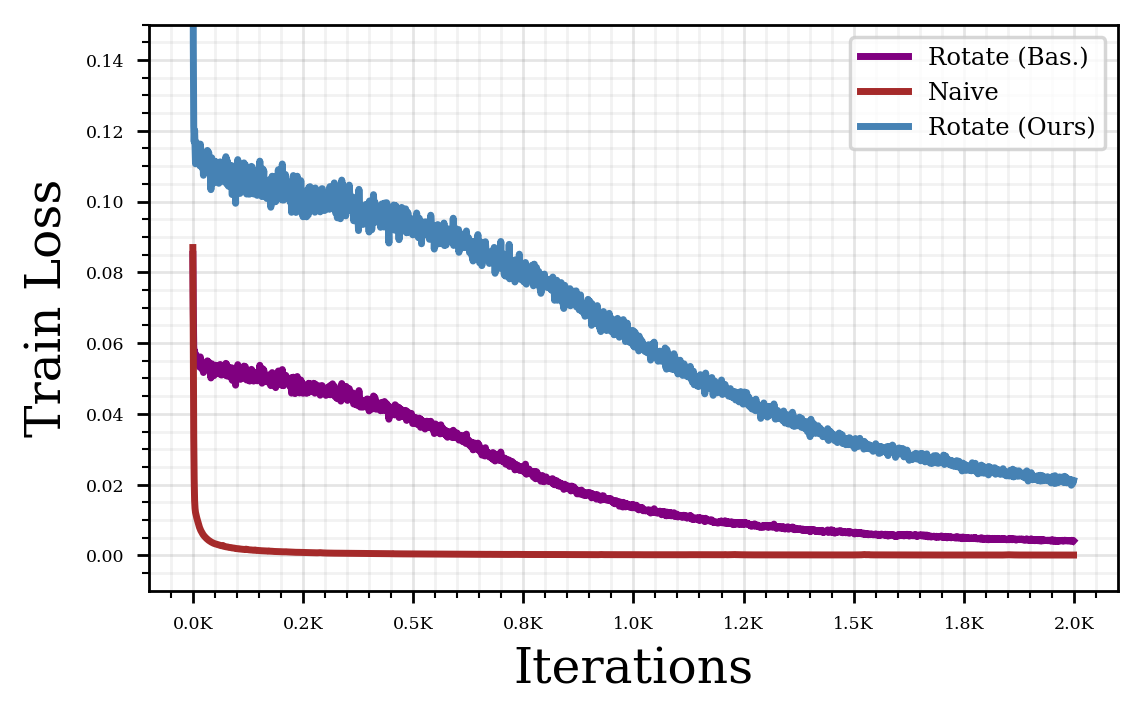}
    \end{minipage}
    \begin{minipage}{.49\linewidth}
        \centering
        \includegraphics[width=\linewidth]{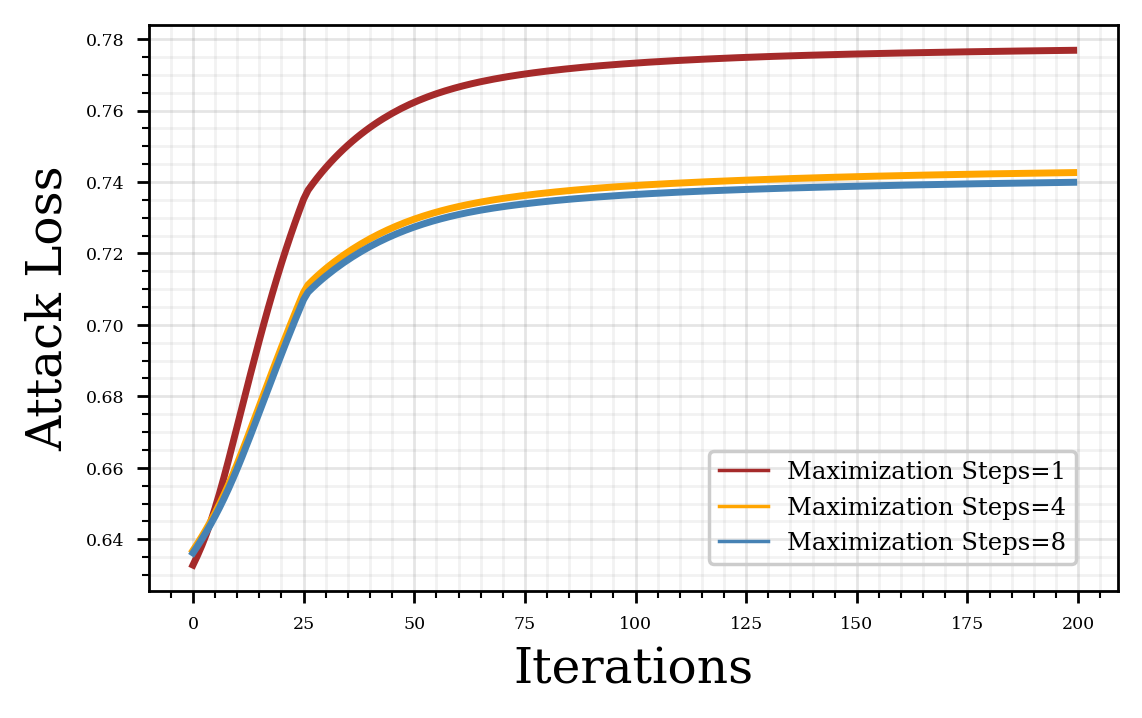}
    \end{minipage}
\end{minipage}
% \vspace{-1.5em}
\begin{minipage}{0.30\linewidth}
    \caption{\small Comparing effect of Adversarial Pixel Restoration Prior on the training loss of surrogate models(\emph{left}) and effect of increasing maximization steps during training on the attack loss(\emph{left}).}
\label{fig: comparing_with_naive_baseline_and_maximization_steps}
\end{minipage}
\vspace{-1em}
\end{figure}

\noindent \textbf{Effect of Adversarial Pixel Restoration Prior:} 
The number of parameters of the surrogate model are significantly higher than the number of input samples, a simple objective of reconstructing the original image can lead to identical mapping. While adding pixel transformations (such as rotation or jigsaw) somewhat alleviates this problem, our adversarial denoising with pixel transformation further resolves it by lowering overfitting, resulting in better generalizability. Fig. \ref{fig: comparing_with_naive_baseline_and_maximization_steps} \emph{(left)} shows training loss comparison between simple reconstruction objective (\emph{Naive}), \cite{li2020practical} and our method. The surrogate auto-encoder quickly collapses to identity as the training loss overfits early on in training without using pixel restoration as prior. On the other hand, our approach allows meaningful representation learning with higher iterations. 
% This is further reflected by higher transferability rates of our adversarial features (Fig.  (\emph{left plot})).

\begin{figure}[!t]
\begin{minipage}{\linewidth}
    \centering
     % lelf lower right up 
    \includegraphics[  trim= 20mm 14mm 20mm 55mm, clip, width=0.32\textwidth]{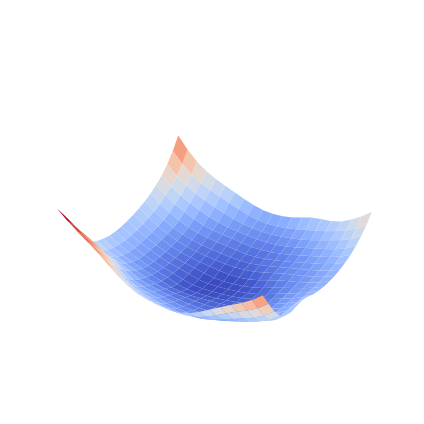}
    \includegraphics[  trim= 20mm 14mm 20mm 55mm, clip, width=0.32\textwidth]{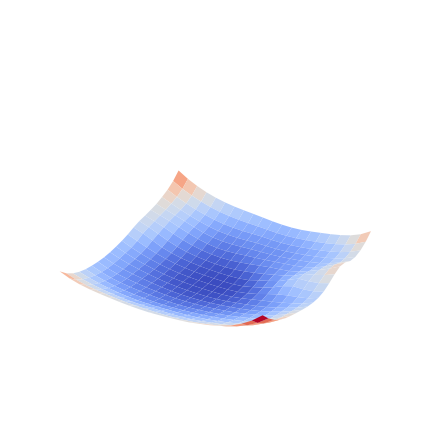}
    \includegraphics[  trim= 20mm 14mm 20mm 55mm, clip, width=0.32\textwidth]{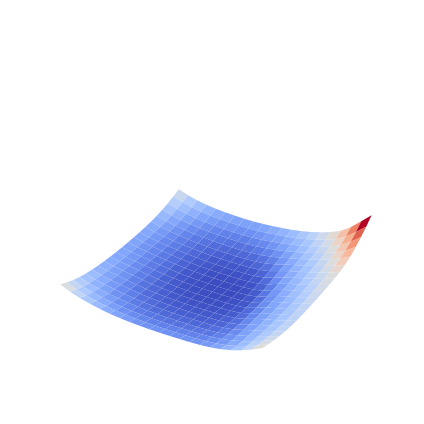}\\
    \vspace{-1em}
    \includegraphics[ trim= 20mm 12mm 20mm 55mm, clip, width=0.32\textwidth]{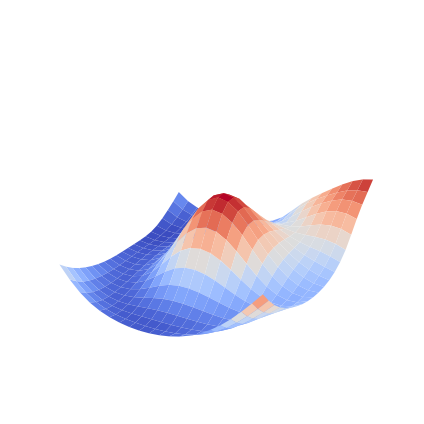}
    \includegraphics[ trim= 20mm 12mm 20mm 55mm, clip, width=0.32\textwidth]{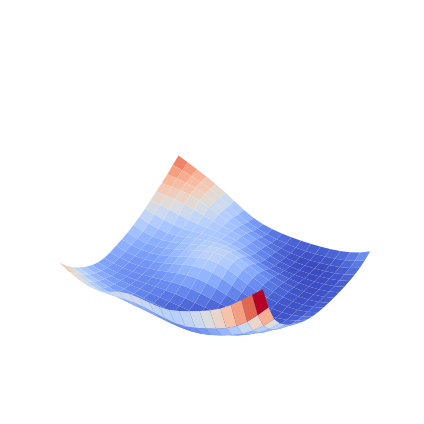}
    \includegraphics[ trim= 20mm 12mm 20mm 55mm, clip, width=0.32\textwidth]{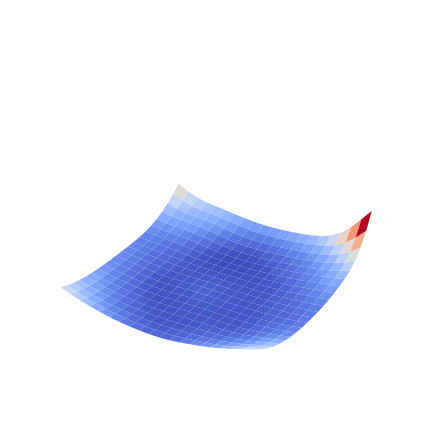}
\end{minipage}
\hspace{0.01\linewidth}
\vspace{-2em}
    \caption{\small Loss landscape of surrogate models with increasing perceptual budget from left to right.The first row shows the loss surface on the clean samples, while as the second row plots the loss surface with respect to the corresponding adversaries.}
    \label{Loss_surface_varying_robustness}
\vspace{-1em}
\end{figure}

\begin{figure*}[!t]
\begin{minipage}{0.5\textwidth}
  
\centering
\begin{minipage}{0.24\textwidth}
      \centering
       % lelf lower right up trim= 7.5mm 0mm 0mm 10mm
    \includegraphics[height=1.2cm, width=\linewidth]{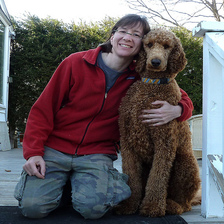}
  \end{minipage}
  \begin{minipage}{0.24\textwidth}
      \centering
       % lelf lower right up trim= 7.5mm 0mm 0mm 10mm
   \includegraphics[height=1.2cm, width=\linewidth]{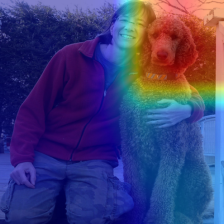}
  \end{minipage}
\begin{minipage}{0.24\textwidth}
      \centering
       % lelf lower right up
  \includegraphics[height=1.2cm, width=\linewidth]{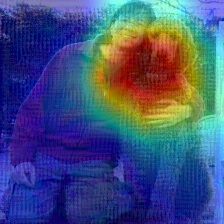}
  \end{minipage}
     \begin{minipage}{0.24\textwidth}
      \centering
       % lelf lower right up trim= 7.5mm 0mm 0mm 10mm
    \includegraphics[height=1.2cm, width=\linewidth]{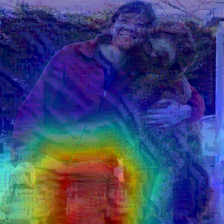}
  \end{minipage}
  \\
  \begin{minipage}{0.24\textwidth}
      \centering
       % lelf lower right up trim= 7.5mm 0mm 0mm 10mm
    \includegraphics[height=1.2cm, width=\linewidth]{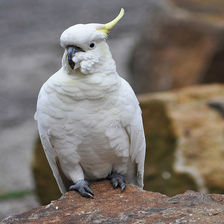}
  \end{minipage}
  \begin{minipage}{0.24\textwidth}
      \centering
       % lelf lower right up trim= 7.5mm 0mm 0mm 10mm
    \includegraphics[height=1.2cm, width=\linewidth]{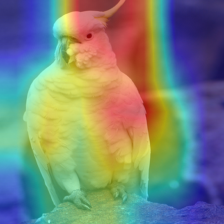}
  \end{minipage}
  \begin{minipage}{0.24\textwidth}
      \centering
       % lelf lower right up trim= 7.5mm 0mm 0mm 10mm
    \includegraphics[height=1.2cm, width=\linewidth]{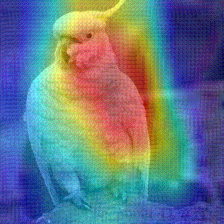}
  \end{minipage}
    \begin{minipage}{0.24\textwidth}
      \centering
       % lelf lower right up trim= 7.5mm 0mm 0mm 10mm
    \includegraphics[height=1.2cm, width=\linewidth]{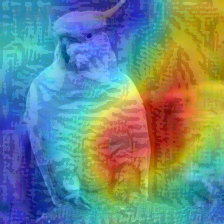}
  \end{minipage}
  \\
  \begin{minipage}{0.24\textwidth}
      \centering
       % lelf lower right up trim= 7.5mm 0mm 0mm 10mm
     \includegraphics[height=1.2cm, width=\linewidth]{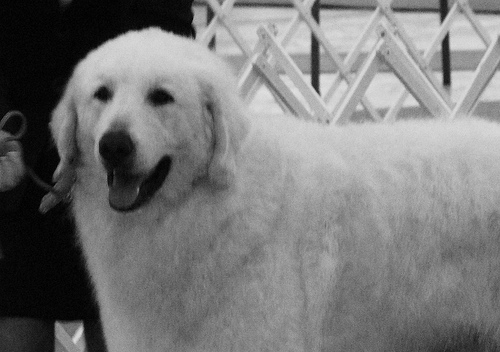}
     \footnotesize Samples
  \end{minipage}
      \begin{minipage}{0.24\textwidth}
      \centering
       % lelf lower right up
    \includegraphics[height=1.2cm, width=\linewidth]{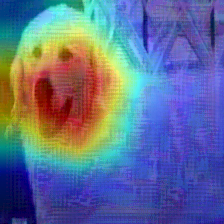}
    \footnotesize Clean Maps
  \end{minipage}
  \begin{minipage}{0.24\textwidth}
      \centering
       % lelf lower right up trim= 7.5mm 0mm 0mm 10mm
    \includegraphics[height=1.2cm, width=\linewidth]{images/attn_vis/baseline/4.png}
    \footnotesize \cite{li2020practical}
  \end{minipage}
   \begin{minipage}{0.24\textwidth}
      \centering
       % lelf lower right up
    \includegraphics[height=1.2cm, width=\linewidth]{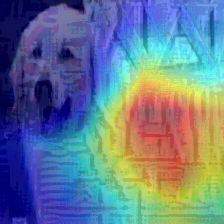}
    \footnotesize Ours
  \end{minipage}

  \end{minipage}
  \hfill
  \begin{minipage}{0.48\textwidth}
  \vspace{-1em}
    \caption{ \small GradCAM \cite{selvaraju2017grad} explanation of adversarial examples. The activation maps were generated on ImageNet pretrained ResNet50 model. Adversarial examples are transferred from surrogate  trained via our adversarial rotation restoration task at the perceptual budget of $\epsilon \le 0.1$ \cite{li2020practical}. Our approach significantly shifts the attention of the model, boosting the mis-classifcation rates on the adversarial examples (see Appendix \ref{appendix: attention dispersion} 
    for more qualitative examples). }
  \label{fig:attn_vis}
  \end{minipage}
  \vspace{-1em}
\end{figure*}

\noindent \textbf{Effect of Perceptual Budget ($\epsilon$):} Our adversarial training (Algo. \ref{algo: Adversarial Denoising and Restoration}) is computationally efficient as it is based on a single attack (maximization) step. The effect of attack step size on model generalizability and hence its transferability is presented in Table \ref{tab: effect of epsilon}. We observe that adversarial perturbation computed on models trained with a smaller $\epsilon$ results in significant improvements in terms of attack transferability. In Fig. \ref{Loss_surface_varying_robustness}, we plot the loss surface around clean and adversarial examples on surrogate models trained with increasing perceptual budget. While the loss surface around clean examples becomes smoother as perceptual budget increases, it becomes harder to maximize the reconstruction error or flip decisions on the excessively smooth loss surface during attack. This behaviour is in line with \cite{springer2021little} which shows that slightly robust classifier models generate highly transferable adversarial examples. This is further evident by the shift in attention caused by our method (Fig. \ref{fig:attn_vis}).

\noindent\textbf{Effect of Iterative Attack:} During our adversarial training, increasing the attack iterations does not help to further boost the adversarial transferability; rather, performance often degrades significantly. In Fig. \ref{fig: comparing_with_naive_baseline_and_maximization_steps} \emph{(right)}, attack loss is plotted for surrogate models (at pereceptual budget $\epsilon=4$) trained with different number of attack iteration (or maximization steps).As the surrogate model becomes more robust with increasing attack iterations, we notice that maximizing the attack objective becomes more challenging.

\noindent \textbf{Contribution of Losses:} We explore the effect of our proposed pixel and feature reconstruction losses (Eq. \ref{eq: overall_loss}) in Table \ref{tab: Loss_contrb}. We observe that feature reconstruction compliments the pixel reconstruction and leads to better surrogate model with more transferable adversarial space.
\begin{table}[!t]
    \begin{minipage}{0.49\textwidth}
        \centering
        \scalebox{0.75}{
        \begin{tabular}{l|c|c|c}
        \toprule 
            \rowcolor{Gray}
          {Perceptual Budget $\epsilon$ ($\rightarrow$)} & {$\frac{2}{255}$}& {$\frac{4}{255}$} & {$\frac{8}{255}$}\\
        \midrule
        
        Rotation & 26.87 & 27.86 & 34.01 \\
        \midrule
        Jigsaw & 24.81 & 27.77 & 35.94 \\ 
        \midrule
        Prototypical & 22.84 & 22.30 & 24.04\\ 
        \bottomrule
        \end{tabular}}
        \vspace{0.5em}
    \caption{\small Effect of Perceptual Budget ($\epsilon$). Top-1 average accuracy is reported on convolution networks under limited data constraint (Table \ref{tab:no-box table}). }
    \label{tab: effect of epsilon}
\end{minipage}
\hfill
\begin{minipage}{0.49\textwidth}
        \small \centering
        \scalebox{0.75}{
        \begin{tabular}{c|c|c}
        \toprule 
            \rowcolor{Gray}
          Training Loss ($\rightarrow$) & $\mathcal{L}_{out}$ & $\mathcal{L}_{out} +  \mathcal{L}_{feature}$\\
        \midrule
        
        Rotation &  29.62 & \textbf{26.87}\dec{3.03} \\
        \midrule
        Jigsaw &  27.13 & \textbf{24.81}\dec{4.59} \\ 
        \midrule
        Prototypical &  23.98 & \textbf{22.84}\dec{0.84}\\ 
        \bottomrule
        \end{tabular}
        }
        \vspace{0.5em}
        \caption{\small Contribution of Loss components. Top-1 average accuracy is reported on convolution networks under limited data constraint (Table \ref{tab:no-box table}).  }
        \label{tab: Loss_contrb}
\end{minipage}
\vspace{-2em}
\end{table} 

\begin{figure}[!t]
\small \centering
\begin{minipage}{0.68\textwidth}
    \begin{minipage}{.49\textwidth}
        \centering
        \includegraphics[width=\linewidth]{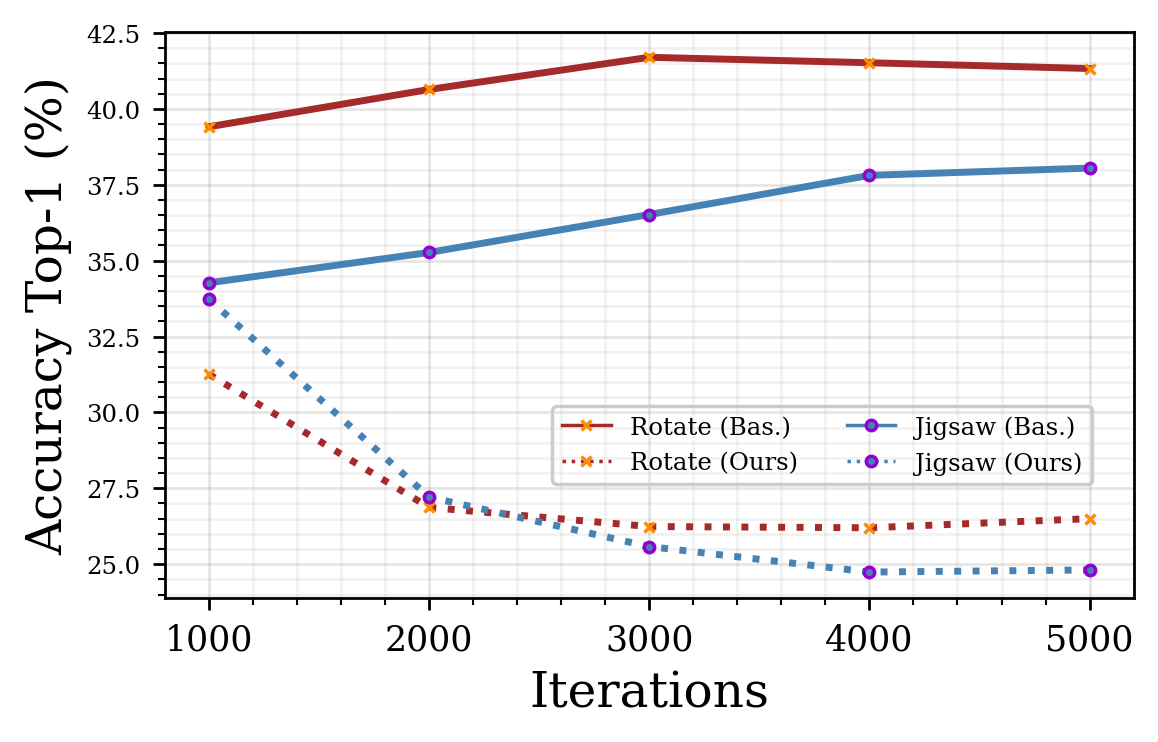}
    \end{minipage}
    \begin{minipage}{.49\textwidth}
        \centering
        \includegraphics[width=\linewidth]{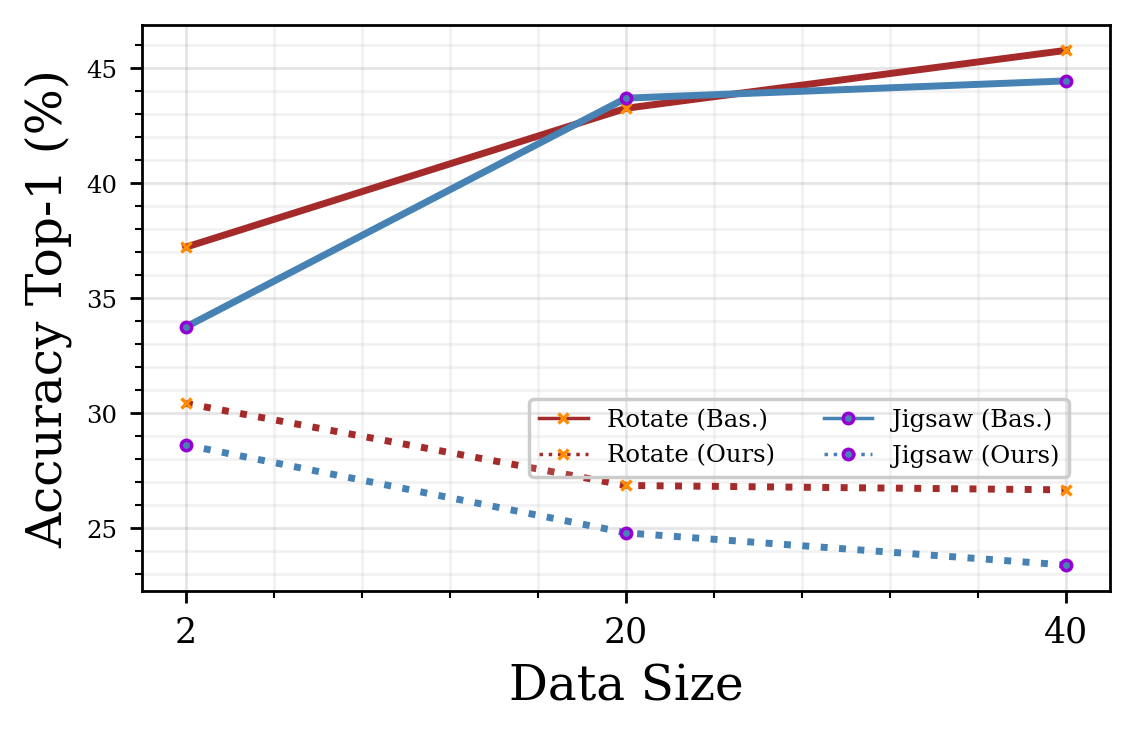} 
	\end{minipage}
\end{minipage}
\begin{minipage}{0.30\textwidth}
\vspace{-1em}
    \caption{\small Top-1 (\%) average accuracy against convolutional networks. The performance of our method improves with more training iterations and data size in contrast to the baseline \cite{li2020practical}.}
\label{fig: ablation iterations and datasize}
\end{minipage}
\vspace{-0.5em}
\end{figure}

\noindent \textbf{Training iterations and Data Size:}  We explore the effect on transferability of surrogate models w.r.t \textbf{a)} training iterations, and \textbf{b)} the number of data samples in Fig. \ref{fig: ablation iterations and datasize}. Note that \cite{li2020practical} trains a single autoencoder on 20 samples and thus needs 250 autoencoder to attack 5k ImageNet validation samples. The number of autoencoders increases to 2.5k when only 2 samples are available for training of single model. In the same setting, the performance of our method improves with more iterations (Fig. \ref{fig: ablation iterations and datasize} \emph{left plot}) in contrast to \cite{li2020practical}. We report average Top-1 (\%) accuracy across all convolutional models (\emph{lower is better}). Similarly, as we increase the number of data samples during training and reduce the number of autoencoders, the performance of our approach significantly improves as compared to the baseline \cite{li2020practical}. This indicates that our adversarial objective successfully reduces overfitting while increases generalizability of crafted adversarial perturbations.

\section{Conclusion}
In this work, we show the benefits of adversarial training to learn transferable adversarial perturbations. Our approach trains an effective surrogate by learning to restore adversarial pixel transformations created via our proposed attack. Our adversarial training reduces overfitting during training and can exploit very few data samples to learn meaningful adversarial features while it can also scale to large unsupervised datasets. Our attack is task independent and allows cross-domain attacks (\eg, learning surrogate on comics and transferring its perturbations to models trained on natural images). Our results bring attention to the use of self-supervised adversarial training for transferable adversarial attacks.

\bibliography{egbib}

\newpage
\appendix

\noindent\begin{huge} \textbf{Appendix} \vspace{4mm} \end{huge}

We provide transferability results of our self-supervised adversarial perturbations computed at lower perceptual budget ($\ell_\infty \le 0.08$)  for in-domain (Tables  \ref{tab:no-box table eps0.08} \& \ref{tab:no-box table_transformers eps0.08}) and cross-domain (Table \ref{tab: single_surrogate_classification_results_eps0.08}) settings in Appendix \ref{appendix: results_eps_0.08}. Our \emph{Adversarial Pixel Restoration} approach remains effective as compared to the baseline \cite{li2020practical} in fooling the Convolutional Networks, Vision Transformers as well as  state-of-the-art input processing defense \cite{naseer2020self} (Fig. \ref{fig: Adversarial purification nobox} \& \ref{fig: Adversarial purification single model} in Appendix \ref{appendix: adv purif}). In appendix \ref{adv_examples} and \ref{appendix: attention dispersion}, we visualize adversarial examples and analyze the attention shift cause by our attack, respectively. Finally,  in appendix \ref{appendix: pseudocode}, we provide psuedocode for self-supervised adversarial training of surrogate models using our approach.

\section{Adversarial Transferability under $\ell_\infty \le 0.08$}
 \label{appendix: results_eps_0.08}

\begin{table}[h]
\centering\small
  \scalebox{0.8}[0.8]{
    \begin{tabular}{lccccccccl}
        \toprule
        \rowcolor{Gray} 
        Transformation & Method  & VGG-19  & Inc-V3  & Res152  & Dense121  &  SeNet & WRN  &  MNet-V2  & Average \\ \midrule
        \multirow{2}{*}{\rotatebox[origin=c]{0}{Jigsaw}} & \cite{li2020practical}  & 40.00 &  58.20 &  55.66 & 50.30 & 66.62& 59.52  & 34.60 & 52.13 \\ 
         & Ours   & 30.88 &  37.82 & 46.14 & 38.04 & 52.18  & 42.62 & 26.32 & \textbf{39.14}\dec{12.99}\\ 
         \midrule
        \multirow{2}{*}{\rotatebox[origin=c]{0}{Rotation}} & \cite{li2020practical}   & 38.88 &  56.16 & 57.06  & 49.56 & 65.30& 58.14 & 34.64 & 51.39\\
        & Ours   & 33.48 & 37.78  & 47.16 & 38.98 & 52.96  & 43.74 & 28.52 & \textbf{40.37}\dec{11.02}\\
        \midrule
        \multirow{2}{*}{\rotatebox[origin=c]{0}{Prototypical}}  & \cite{li2020practical} & 30.08 &  45.74& 47.28 & 37.66 & 54.42 & 44.82  & 27.32 & 41.05\\
        &  Ours 	 &30.44  & 31.96  & 42.76 & 34.50 & 49.20  & 38.80 & 23.94& \textbf{35.94}\dec{5.11}
        \\\midrule
        % \multicolumn{9}{l}{\footnotesize{$^\ast$ }}\\
    \end{tabular}}
    \vspace{0.3em}
    \caption{Comparative analysis of adversarial transferability. Results (top-1 (\%), \emph{lower is better}) are reported on 5k images from ImageNet validation set under the perceptual budget of $\epsilon \le 0.08$. Our attack provides favorable results as compared to \cite{li2020practical}.}
    \label{tab:no-box table  eps0.08}
\end{table}

\begin{table}[h]
    \begin{minipage}{.55\textwidth}
        \centering\small
        \setlength{\tabcolsep}{5pt}
        
        \scalebox{0.8}[0.8]{
    \begin{tabular}{lcccccl}
        \toprule
        \rowcolor{Gray} 
         Transformation & Method  & Deit-T  & Deit-S  & ViT-T  & ViT-S  & Average \\ \midrule
        \multirow{2}{*}{\rotatebox[origin=c]{0}{Jigsaw}} & \cite{li2020practical}  & 51.32 &  68.26 &  46.68 & 68.16 & 58.61 \\ 
         & Ours   & 53.48 &  67.50 & 32.78 & 62.60 & \textbf{54.09}\dec{4.52}\\ 
         \midrule
        \multirow{2}{*}{\rotatebox[origin=c]{0}{Rotation}} & \cite{li2020practical}   & 53.22&  68.36 & 47.94  & 66.72 & 59.06\\
        & Ours   & 51.16 & 66.10  & 32.68 & 60.20 & \textbf{52.54}\dec{6.52}\\
        \midrule
        \multirow{2}{*}{\rotatebox[origin=c]{0}{Prototypical}}  & \cite{li2020practical} & 47.44 &  64.22& 32.94 & 60.06 & 51.17\\
        &  Ours 	 &45.70  & 62.26 & 29.50 & 57.98 & \textbf{48.86}\dec{2.31}  
        \\\midrule
        % \multicolumn{9}{l}{\footnotesize{$^\ast$ }}\\
    \end{tabular}}
    \end{minipage}
    \hfill
   \begin{minipage}{.34\textwidth}
   \vspace{-1em}
    \caption{\small Comparative analysis of adversarial transferability for Vision  Transformers on 5k images \cite{li2020practical} from ImageNet validation set under the perceptual budget of $\epsilon \le 0.08$. ViTs are relatively more robust as compared to CNNs.}
    \label{tab:no-box table_transformers eps0.08}
    \end{minipage} 
    \vspace{-0.5em}
\end{table}

\begin{table}[h]
    % \begin{minipage}{.44\textwidth}
        \centering\small
        \setlength{\tabcolsep}{6pt}
        
        \scalebox{0.8}[0.8]{
        \begin{tabular}{clca|ca}
        \toprule 
            \rowcolor{Gray}
             & & \multicolumn{2}{c}{Convolutional Networks} & \multicolumn{2}{c}{Vision Transformers} \\ 
             \cline{3-6}
             \rowcolor{Gray}
          Transformation & Method ($\rightarrow$) & {\cite{li2020practical}}& {Ours} &  {\cite{li2020practical}} & Ours  \\
        \midrule
        
        \multirow{3}{*}{\rotatebox[origin=c]{0}{Rotation}} & CoCo & 38.47 & \textbf{35.72} & 48.24 & \textbf{46.44} \\
        & Paintings & 36.81 & \textbf{30.48 }  & 45.92 & \textbf{42.04 } \\ 
        & Comics & 62.46 & \textbf{39.16 } & 70.39 & \textbf{53.03 } \\ 
        \midrule
        \multirow{3}{*}{\rotatebox[origin=c]{0}{Jigsaw}} & CoCo & 51.75 & \textbf{43.6} & 61.95 & \textbf{55.36}\\
        & Paintings&  51.56 & \textbf{44.42} & 61.59 & \textbf{56.64}\\ 
        & Comics & 69.89 & \textbf{52.65 }  & 75.66 & \textbf{62.15 }  \\ 
        \bottomrule
        \end{tabular}
        }
    % \end{minipage}
    % \hfill
    \vspace{0.3em}
%   \begin{minipage}{.55\textwidth}
    \caption{\small Adversarial perturbations (\emph{$\epsilon = 0.08$}) are transferred from a  single surrogate auto-encoder trained on CoCo, Paintings or Comics datasets to different Convolutional and Transformer based models. Results ( top-1 (\%) accuracy, \emph{lower is better}) are averaged across all models. Details on these models are provide in Sec. \ref{sec: Experimental Protocols}}
    \label{tab: single_surrogate_classification_results_eps0.08}
    % \end{minipage} 
\end{table}

\subsection{Adversarial Transferability against Neural Purification Defense }
\label{appendix: adv purif}

\begin{figure}[!htb]
\small \centering
\begin{minipage}{0.8\textwidth}
    \begin{minipage}{.49\textwidth}
        \centering
        \includegraphics[width=\linewidth]{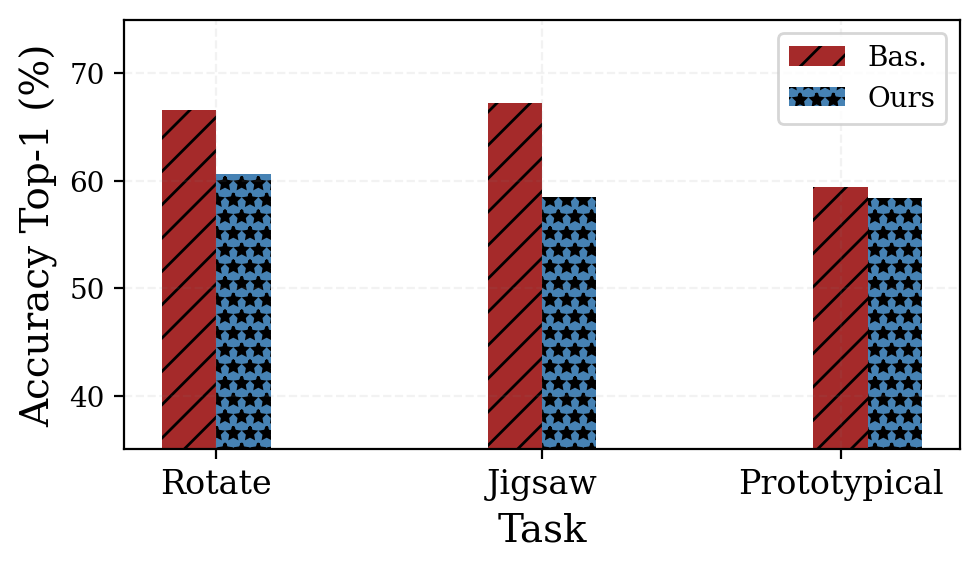}
    \end{minipage}
    \begin{minipage}{.49\textwidth}
        \centering
        \includegraphics[width=\linewidth]{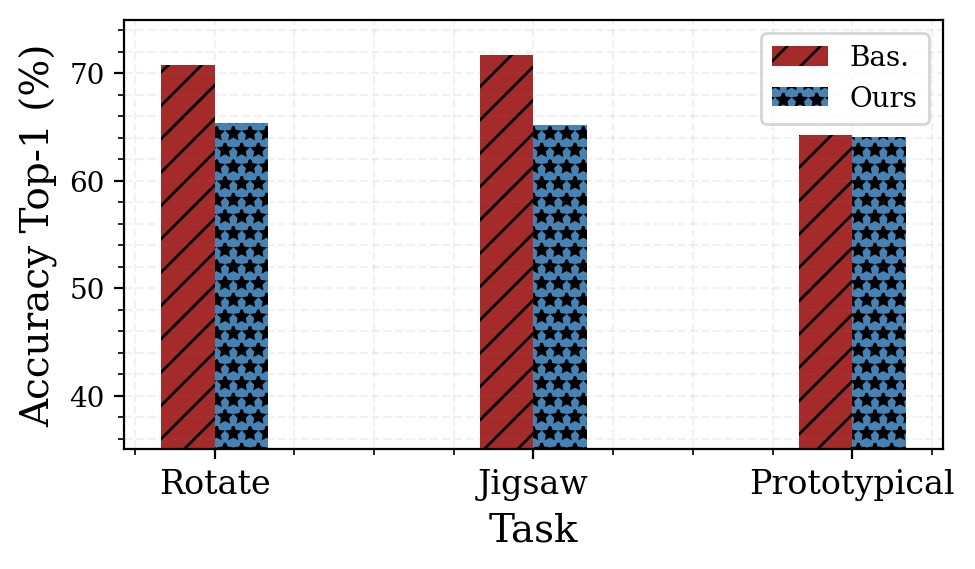} 
	\end{minipage}
\end{minipage}
% \begin{minipage}{0.30\textwidth}
% \vspace{-1em}
    \caption{\small We test the vulnerability of the state-of-the-art input processing defense NRP \cite{naseer2020self} against our self-supervised attack. Adversarial perturbations are transferred from auto-encoders under the constraint of limited in-domain samples \cite{li2020practical} to Convolutional Networks (\emph{left plot}), and Vision Transformers (\emph{right plot}) protected by NRP defense. We report Top-1 (\%) (\emph{lower is better}) averaged across the models. Our method consistently improves the attack success rate. }
\label{fig: Adversarial purification nobox}
% \end{minipage}
\end{figure}

\begin{figure}[h]
\small \centering
\begin{minipage}{0.8\textwidth}
    \begin{minipage}{.49\textwidth}
        \centering
        \includegraphics[width=\linewidth]{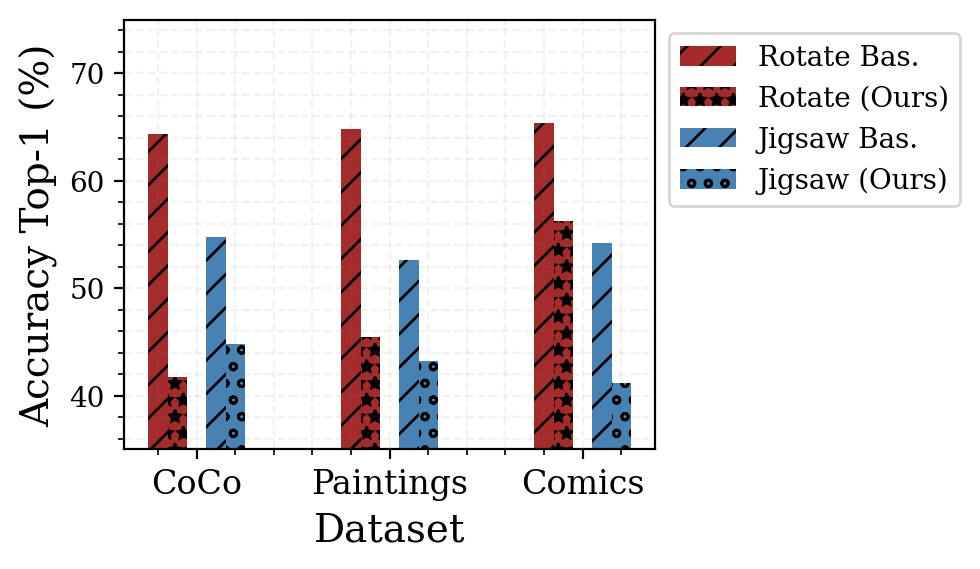}
    \end{minipage}
    \begin{minipage}{.49\textwidth}
        \centering
        \includegraphics[width=\linewidth]{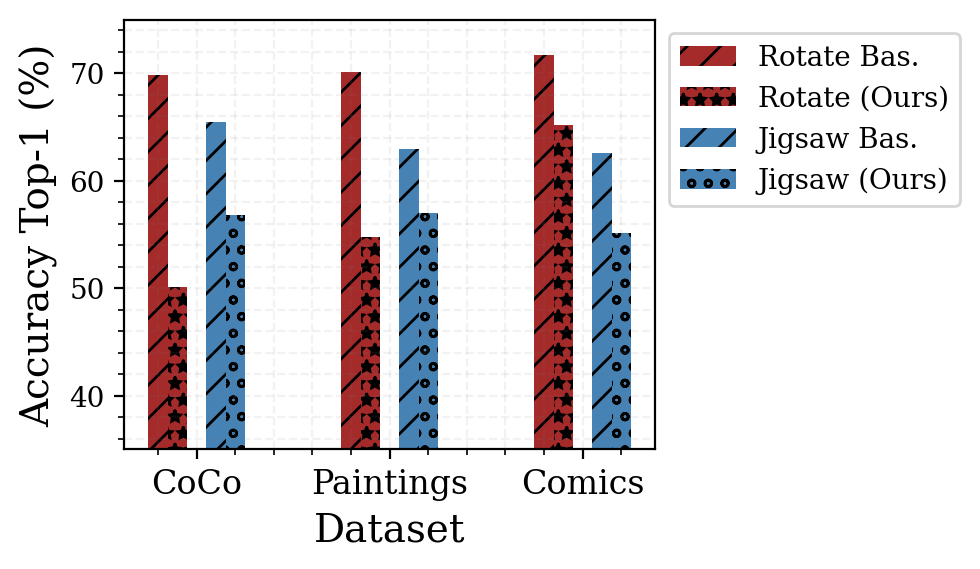} 
	\end{minipage}
\end{minipage}
% \begin{minipage}{0.30\textwidth}
% \vspace{-1em}
    \caption{\small   We test the vulnerability of the state-of-the-art input processing defense NRP \cite{naseer2020self} against our self-supervised attack under cross-domain setting (Sec. \ref{sec: Experimental Protocols}). Adversarial perturbations are transferred from single auto-encoder trained on CoCo, Paintings or Comics to Convolutional Networks (\emph{left plot}), and Vision Transformers (\emph{right plot}) protected by NRP defense. We report Top-1 (\%) (\emph{lower is better}) averaged across the models. Our method consistently improves the attack success rate.}
\label{fig: Adversarial purification single model}
% \end{minipage}
\end{figure}

\section{Adversarial Examples}\label{adv_examples}

\begin{figure}[!h]
\begin{minipage}{\linewidth}
    \centering
     % lelf lower right up 
    \includegraphics[width=0.12\textwidth, height=1.2cm]{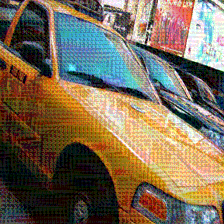}
    \includegraphics[width=0.12\textwidth, height=1.2cm]{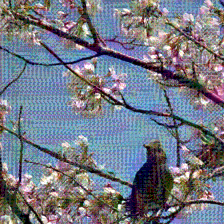}
    \includegraphics[width=0.12\textwidth, height=1.2cm]{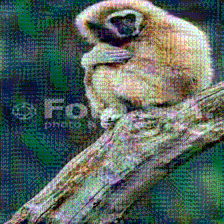}
    \includegraphics[width=0.12\textwidth, height=1.2cm]{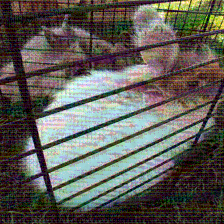}
    \includegraphics[width=0.12\textwidth, height=1.2cm]{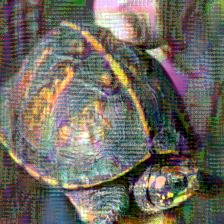}
    \includegraphics[width=0.12\textwidth, height=1.2cm]{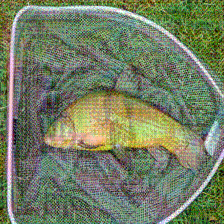}
    \\
    \vspace{0.05em}
     \includegraphics[width=0.12\textwidth, height=1.2cm]{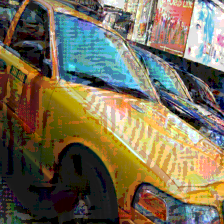}
    \includegraphics[width=0.12\textwidth, height=1.2cm]{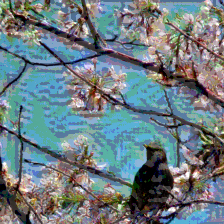}
    \includegraphics[width=0.12\textwidth, height=1.2cm]{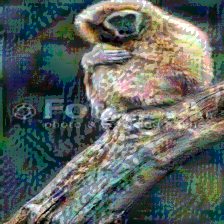}
    \includegraphics[width=0.12\textwidth, height=1.2cm]{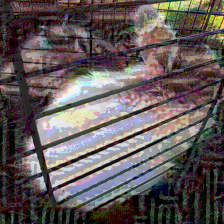}
    \includegraphics[width=0.12\textwidth, height=1.2cm]{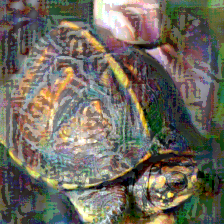}
    \includegraphics[width=0.12\textwidth, height=1.2cm]{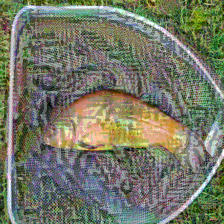}
\end{minipage}
\hspace{0.01\linewidth}
    \caption{\small Adversarial examples crafted on surrogate models trained on limited data, with perturbation bound $\epsilon \leq 0.1$. The top and bottom row show adversarial examples crafted on surrogate model trained using \cite{li2020practical} and ours method, respectively. }
    \label{adversarial_examples}
    \vspace{1cm}
\end{figure}

\section{Attention Dispersion}
\label{appendix: attention dispersion}

\begin{figure*}[!h]
\begin{minipage}{0.5\textwidth}
  
\centering
\begin{minipage}{0.24\textwidth}
      \centering
       % lelf lower right up trim= 7.5mm 0mm 0mm 10mm
    \includegraphics[height=1.2cm, width=\linewidth]{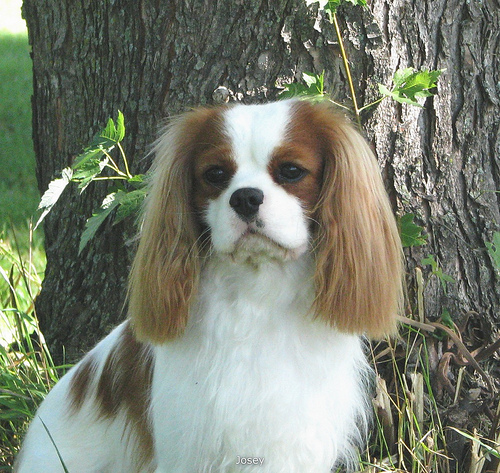}
  \end{minipage}
  \begin{minipage}{0.24\textwidth}
      \centering
       % lelf lower right up trim= 7.5mm 0mm 0mm 10mm
   \includegraphics[height=1.2cm, width=\linewidth]{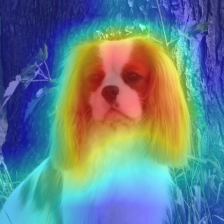}
  \end{minipage}
\begin{minipage}{0.24\textwidth}
      \centering
       % lelf lower right up
  \includegraphics[height=1.2cm, width=\linewidth]{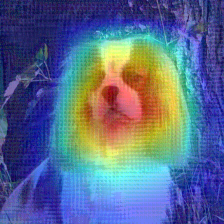}
  \end{minipage}
     \begin{minipage}{0.24\textwidth}
      \centering
       % lelf lower right up trim= 7.5mm 0mm 0mm 10mm
    \includegraphics[height=1.2cm, width=\linewidth]{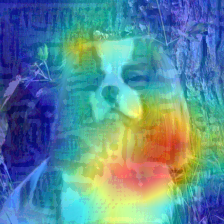}
  \end{minipage}
  \\
  \begin{minipage}{0.24\textwidth}
      \centering
       % lelf lower right up trim= 7.5mm 0mm 0mm 10mm
    \includegraphics[height=1.2cm, width=\linewidth]{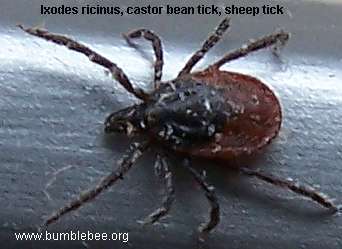}
  \end{minipage}
  \begin{minipage}{0.24\textwidth}
      \centering
       % lelf lower right up trim= 7.5mm 0mm 0mm 10mm
    \includegraphics[height=1.2cm, width=\linewidth]{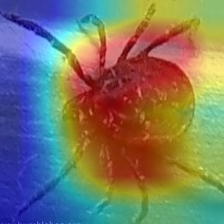}
  \end{minipage}
  \begin{minipage}{0.24\textwidth}
      \centering
       % lelf lower right up trim= 7.5mm 0mm 0mm 10mm
    \includegraphics[height=1.2cm, width=\linewidth]{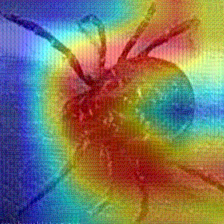}
  \end{minipage}
    \begin{minipage}{0.24\textwidth}
      \centering
       % lelf lower right up trim= 7.5mm 0mm 0mm 10mm
    \includegraphics[height=1.2cm, width=\linewidth]{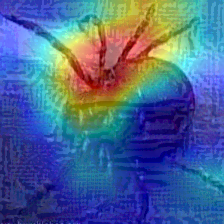}
  \end{minipage}
  \\
  \begin{minipage}{0.24\textwidth}
      \centering
       % lelf lower right up trim= 7.5mm 0mm 0mm 10mm
     \includegraphics[height=1.2cm, width=\linewidth]{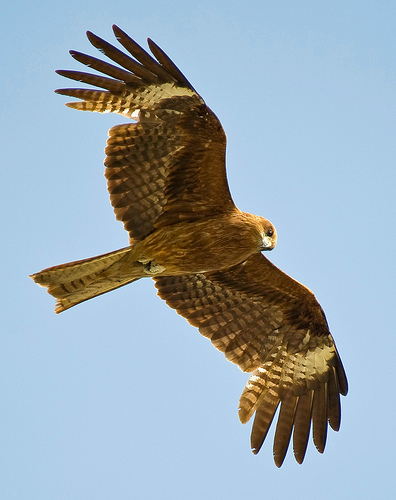}
  \end{minipage}
      \begin{minipage}{0.24\textwidth}
      \centering
       % lelf lower right up
    \includegraphics[height=1.2cm, width=\linewidth]{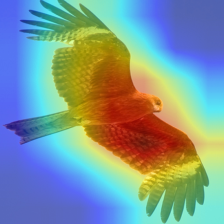}
  \end{minipage}
  \begin{minipage}{0.24\textwidth}
      \centering
       % lelf lower right up trim= 7.5mm 0mm 0mm 10mm
    \includegraphics[height=1.2cm, width=\linewidth]{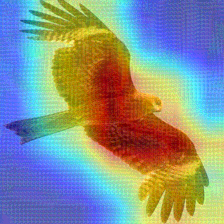}
  \end{minipage}
   \begin{minipage}{0.24\textwidth}
      \centering
       % lelf lower right up
    \includegraphics[height=1.2cm, width=\linewidth]{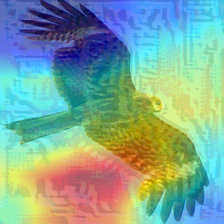}
  \end{minipage}
    \\
      \begin{minipage}{0.24\textwidth}
      \centering
       % lelf lower right up trim= 7.5mm 0mm 0mm 10mm
     \includegraphics[height=1.2cm, width=\linewidth]{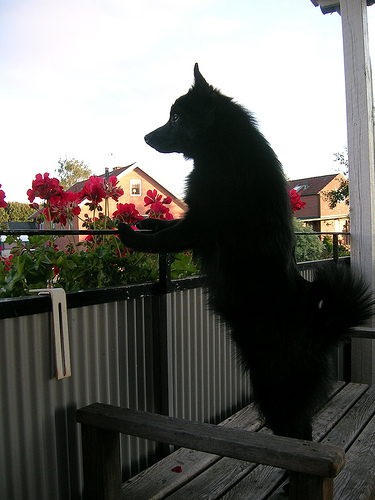}
     \footnotesize Samples
  \end{minipage}
      \begin{minipage}{0.24\textwidth}
      \centering
       % lelf lower right up
    \includegraphics[height=1.2cm, width=\linewidth]{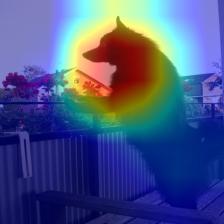}
    \footnotesize Clean Maps
  \end{minipage}
  \begin{minipage}{0.24\textwidth}
      \centering
       % lelf lower right up trim= 7.5mm 0mm 0mm 10mm
    \includegraphics[height=1.2cm, width=\linewidth]{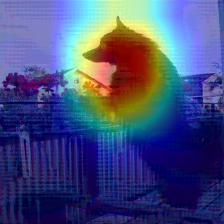}
    \footnotesize \cite{li2020practical}
  \end{minipage}
   \begin{minipage}{0.24\textwidth}
      \centering
       % lelf lower right up
    \includegraphics[height=1.2cm, width=\linewidth]{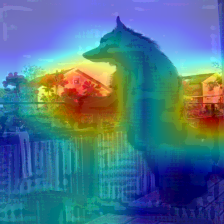}
    \footnotesize Ours
  \end{minipage}
  \end{minipage}
  \hfill
  \begin{minipage}{0.48\textwidth}
  \vspace{-1em}
    \caption{ \small GradCAM \cite{selvaraju2017grad} explanation of adversarial examples. The activation maps were generated on ImageNet pretrained ResNet50 model. Adversarial examples are transferred from surrogate  trained via our adversarial rotation restoration task at the perceptual budget of $\epsilon \le 0.1$ \cite{li2020practical}. Our approach significantly shifts the attention of the model, boosting the mis-classifcation rates on the adversarial examples.}
  \label{fig:attn_vis_appendix}
  \end{minipage}
  
\end{figure*}

\section{Training Algorithm}
\label{appendix: pseudocode}
\begin{algorithm}[!h]
   \caption{Adversarial Pixel Restoration: Pytorch style Pseudocode }
   \label{alg:opl}
   
    \definecolor{codeblue}{rgb}{0.25,0.5,0.5}
    \lstset{
      basicstyle=\fontsize{8.2pt}{8.2pt}\ttfamily\bfseries,
      commentstyle=\fontsize{7.2pt}{7.2pt}\color{codeblue},
      keywordstyle=\fontsize{7.2pt}{7.2pt},
    }
    
% numbers=left,firstnumber=0,stepnumber=1]
\begin{lstlisting}[language=python] 
def train(model, optimizer, images, iterations, fgsm_step):
    """ 
    images:     images shaped (B, C, H, W) 
    model:      autoencoder
    """
    attack = FGSM(model, eps=fgsm_step)
    target_images = images.clone()
    
    for i in range(iterations):
        
        # Generate adversarial images
        transformed_images = transform(images)
        adv_images = attack(transformed_images, target_images)
        
        # Get the encoder features and output
        clean_output, clean_enc_output = model(transformed_images)
        adv_output, adv_enc_output = model(adv_images)
        
        # Compute losses
        clean_loss = nn.MSELoss()(clean_output, target_images)
        adv_loss = nn.MSELoss()(adv_output, target_images)
        feat_loss = nn.MSELoss()(adv_enc_output, clean_enc_output)
        loss = clean_loss + adv_loss + feat_loss
        
        # Update model parameters
        optimizer.zero_grad()
        loss.backward()
        optimizer.step()
\end{lstlisting}
\end{algorithm}
\end{document}